%% file: neurips_2024.tex
\documentclass{article}

\usepackage[preprint]{neurips_2024}

\usepackage[utf8]{inputenc} % allow utf-8 input
\usepackage[T1]{fontenc}    % use 8-bit T1 fonts
\usepackage{hyperref}       % hyperlinks
\usepackage{url}            % simple URL typesetting
\usepackage{booktabs}       % professional-quality tables
\usepackage{amsfonts}       % blackboard math symbols
\usepackage{nicefrac}       % compact symbols for 1/2, etc.
\usepackage{microtype}      % microtypography
\usepackage{xcolor}         % colors
\usepackage{amsmath,amssymb}
\usepackage{adjustbox}
\usepackage{xcolor, colortbl}
\usepackage{wrapfig}
\usepackage{graphicx}
\usepackage{adjustbox}
\usepackage{subcaption}

\DeclareMathOperator*{\argmax}{arg\,max}
\newcommand{\highlight}[1]{\colorbox{blue!10}{#1}}
\newcommand{\oraclehighlight}[1]{\colorbox{red!20}{#1}}

\definecolor{mygray}{gray}{0.4}
\newcommand{\g}[2]{#1\textsubscript{\textcolor{mygray}{$\pm$#2}}}

\hypersetup{
    colorlinks=true,
    citecolor={blue!64!black},
    urlcolor={blue!64!black},
    linkcolor={blue!64!black},
}

\title{Reinforcement Learning for Sequence Design Leveraging Protein Language Models}

\makeatletter
\def\thanks#1{\protected@xdef\@thanks{\@thanks
        \protect\footnotetext{#1}}}
\makeatother

\author{%
  Jithendaraa~Subramanian{
  \normalfont\textsuperscript{1,2 \textdagger}}
  \thanks{
      \textsuperscript{1} McGill University, 
      \textsuperscript{2} Mila -- Quebec AI Institute, 
      \textsuperscript{3} Araya Inc., 
      \textsuperscript{4} BRAC University, 
      \textsuperscript{5} Bangladesh University of Engineering and Technology, \textsuperscript{6} University of Calgary, 
      \textsuperscript{7} CIFAR AI Chair, 
      \textsuperscript{8} Dreamfold
  }
  \thanks{\textdagger~indicates equal contribution}
  \thanks{
    Correspondence: \href{mailto:jithen.subra@gmail.com}{jithen.subra@gmail.com}, \href{mailto:shivakanth.sujit@gmail.com}{shivakanth.sujit@gmail.com}
  }\quad 
  Shivakanth~Sujit{\normalfont\textsuperscript{2,3 \textdagger}}\quad 
  Niloy~Irtisam{\normalfont\textsuperscript{4}}\quad
  Umong~Sain{\normalfont\textsuperscript{5}} \\
  \textbf{Riashat~Islam{\normalfont\textsuperscript{8}}}\quad
  \textbf{Derek~Nowrouzezahrai{\normalfont\textsuperscript{1,2,7}}}\quad 
  \textbf{Samira~Ebrahimi~Kahou{\normalfont\textsuperscript{2,6,7}}}\\ 
}

\begin{document}
\maketitle
\begin{abstract}
Protein sequence design, determined by amino acid sequences, are essential to protein engineering problems in drug discovery. Prior approaches have resorted to evolutionary strategies or Monte-Carlo methods for protein design, but often fail to exploit the structure of the combinatorial search space, to generalize to unseen sequences. In the context of discrete black box optimization over large search spaces, learning a mutation policy to generate novel sequences with reinforcement learning is appealing. Recent advances in protein language models (PLMs) trained on large corpora of protein sequences offer a potential solution to this problem by scoring proteins according to their biological plausibility (such as the TM-score). In this work, we propose to use PLMs as a reward function to generate new sequences. Yet the PLM can be computationally expensive to query due to its large size. To this end, we propose an alternative paradigm where optimization can be performed on scores from a smaller proxy model that is periodically finetuned, jointly while learning the mutation policy. We perform extensive experiments on various sequence lengths to benchmark RL-based approaches, and provide comprehensive evaluations along biological plausibility and diversity of the protein. Our experimental results include favorable evaluations of the proposed sequences, along with high diversity scores, demonstrating that RL is a strong candidate for biological sequence design. Finally, we provide a modular open source implementation can be easily integrated in most RL training loops, with support for replacing the reward model with other PLMs, to spur further research in this domain. The code for all experiments is provided in the supplementary material. 
\end{abstract}

\section{Introduction}

% RLAIF: cite Constitutional ai: Harmlessness from ai feedback, 2022

Proteins are indispensable for virtually every biological function, from providing structural integrity to managing immune responses and facilitating molecular transport within organisms. Despite their diverse roles, our current knowledge of proteins remains incomplete, representing merely a fraction of the vast protein landscape. Generating novel protein sequences holds promise to deepen our understanding of cellular mechanisms, physiological responses, disease pathways, and can significantly accelerate drug development~\citep{stanton2022accelerating}.

The goal of \textit{de novo} protein design is to find sequences $x$ that maximize an objective function $f(x)$, scored by an oracle. This is a challenging discrete black box optimization problem over a combinatorial search space, that grows according to $\lvert \mathcal{A} \rvert^L$ where $L$ is the sequence length of the protein, and $\lvert \mathcal{A} \rvert = 20$ is the number of most commonly appearing $\alpha$-amino acids to build proteins, although there are upto 22 amino acids in the genetic code of life~\citep{crick1968origin, koonin2009origin}. Designing proteins in this large search space has been attempted with various evolutionary methods, from directed evolution~\citep{arnold_1998} to adaptive greedy search like AdaLead~\citep{sinai2020adalead}. A major caveat, however, is that these methods are computationally inefficient, find sequences in the same neighborhood, and do not scale well with increasing sequence length. In this regard, learning a sequence generation policy with reinforcement learning (RL)~\citep{Angermueller2020Model-based} offers a promising solution, with the potential to exploit the structure in the search space. Once such a policy has been trained, actions can be sampled and sequences can be generated at a fraction of the training cost and can generalize to unseen sequences, \emph{even in the absence} of the reward model. 

Recently, there has been massive efforts towards building large protein language models (PLMs) ~\citep{jumper2021highly, lin2023evolutionary, lin2022language, NEURIPS2021_f51338d7, abramson2024accurate} to address the protein folding problem, along with the prediction of scores (e.g., template modeling score) that estimate biological plausibility of a protein sequence and its corresponding folded structure. Motivated by recent research on large language models (LLMs) which provides evidence that LLMs can be repurposed as reward models~\citep{rafailov2024direct}, we propose to use PLMs as the reward function. In our work, we instantiate the 3B parameter version of ESMFold~\citep{lin2023evolutionary} as the oracle reward model. We choose ESMFold over other competitive PLMs such as AlphaFold2 due to its favourable inference, offering upto \emph{$60\times$ faster inference time}, which is crucial in an RL loop.

% \jith{Tie up some loose ends and talk about \textbf{why} ESM, "Language models generalize beyond natural proteins" paper to justify why ESM and why PLMs for seq design} \\

Our work takes a sequence first approach, similar to EvoDiff~\citep{evodiff} and LaMBO-2~\citep{gruver2024protein}. While protein structures are what ultimately determine their function, sequence determines structure. Moreover, a structure-first approach for protein design still needs to generate sequences, by solving the \emph{inverse folding problem} in order to successfully synthesize the protein in the laboratory. We address the problem of protein design by learning policies that directly generate protein sequences, by learning to mutate random sequences. Our contributions are as follows:

\begin{itemize}

    \item The primary purpose of this paper is to investigate and benchmark the performance of RL algorithms where the reward is given by an oracle PLM, ESMFold~\citep{lin2023evolutionary} in our case.

    \item Since the oracle reward model is expensive to query, we demonstrate an alternate approach that requires fewer evaluations under the oracle model, called ESM with \textit{Proxy-Finetuning} (ESM-PF). \textit{ESM-PF} jointly learns a proxy reward model \textit{and} the sequence generation policy with results comparable to that of directly optimizing on the oracle. The proxy model is faster to query due to its smaller size, and is trained by periodically finetuning on pairs of previously generated sequences and their oracle scores $(x_i, y_i)$.
    
    \item We perform a multi-objective evaluation of the proposed protein sequences in both the oracle and ESM-PF settings, across various sequence lengths, with consideration to biological plausibility and diversity of the sequences, both in the sequence and structure space.

    \item We provide a modular open source implementation, that provides easy support for both the addition of other exploration algorithms (evolutionary, RL-based, generative model based), as well as replacing the ESMFold reward model with a PLM of the user's choice (such as AlphaFold).
\end{itemize}

\section{Related Work}

Several methods have been explored for designing biological sequences. 

\textbf{Evolutionary Algorithms.} Designing sequences by directed evolution~\citep{arnold_1998} was one of the earliest attempts to solve the protein design problem, wherein random mutations are performed on a set of sequences and the best performing ones are greedily chosen to be mutated further. AdaLead~\citep{sinai2020adalead} proposes selecting the sequences that are within a fraction $k$ of the best scoring sequence for further mutation. The fraction $k$ controls the greediness of the algorithm. Setting the threshold based on the score rather than the number of candidate sequences allows the algorithm to adapt to the optimization landscape; it can select many candidates in regions where the landscape is flat, thereby promoting diversity, while being able to focus on the best sequences when there are prominent peaks in the landscape. Covariance Matrix Adaptation Evolutionary Strategy (CMA-ES)~\citep{hansen2001completely} generates a population of candidate sequences from a set of parents and then selects the top $k$ sequences in a round to serve as parents the next round. The candidates are generated by sampling from a multivariate normal distribution, whose mean and covariance are obtained from the current set of parents, which gets updated at the end of every round. Proximal Exploration (PEX)~\citep{ren2022proximal} modifies the optimization objective to encourage selection of candidates that are close to wild type sequences. This is done by adding a regularization term that measures the edit distance of the sequence to the wild type.

% \riashat{TODO:}
% Evolutionary: PEX~\citep{ren2022proximal}

\textbf{Reinforcement Learning and GFlowNets.} DyNAPPO \citep{Angermueller2020Model-based}, is a model-based reinforcement learning algorithm that generates protein sequences by optimizing a surrogate reward function learned through an ensemble. To promote exploration, the reward function penalizes identical sequences. On the other hand, GFlowNets were introduced as a method to sample from discrete, compositional objects by sequentially constructing the object~\citep{bengio2021flow, bengio2023gflownet}. Rather than the goal of reward maximization as in RL, GFlowNets learn to sample proportional to the reward function. This makes them well-suited for sampling from a diverse set of terminal states. For biological sequence design, GFlowNets \citep{jain2022biological, jain2023multi} autoregressively construct the sequence by selecting one amino acid residue at a time. The methodology borrows some principles from Bayesian optimization, such as the estimation of epistemic uncertainty and the use of an acquisition function to score candidates.

\textbf{Generative Models.} Generative models such as discrete diffusion models, trained on large-scale protein sequence datasets, can generate novel proteins in sequence space~\citep{evodiff}. In contrast, diffusion \citep{se3_diff} and flow-based models on the SE(3) manifold such as FoldFlow~\citep{foldflow} and FoldFlow-2~\citep{huguet2024sequence} can generate novel proteins in structure space but are unable to leverage the extensive sequence datasets. Methods such as RFDiffusion~\citep{watson2023novo} have showed promising results employing a structure-first approach. Design by Adaptive Sampling (DbAS)~\citep{brookes2018design} trains a variational autoencoder~\citep{kingma2013auto} in an unsupervised manner, while accounting for uncertainty in the oracle model. The follow-up work on Conditioning by Adaptive Sampling (CbAS)~\citep{brookes2019conditioning} addresses the pathological behaviour of the oracle -- wherein the oracle provides wildly unreliable scores for sequences in a neighborhood far away from where it was trained -- by encoding knowledge about regions where the oracle is expected to be accurate. There are also other approaches to sequence design based on GANs~\citep{goodfellow2014generative}, such as activation maximization and FB-GAN for generating DNA sequences~\citep{killoran2017generating, gupta2018feedback}.

\textbf{MCMC}. Biological sequences can also be generated by sampling from energy-based models. MCMC and simulated annealing can generate sequences by optimizing energy functions that evaluate compatibility with constraints on the protein structures \citep{hie2022high}, leveraging protein language models for structure prediction. A smoothed energy function can also be learned \citep{frey2023protein}, and sequences can be sampled from the smoothed data manifold using MCMC, following one-step denoising.

\textbf{Bayesian Optimization (BO)} is a standard approach for optimizing black box functions using a surrogate model of the objective~\citep{movckus1975bayesian, snoek2012practical} and typically involves maximizing an acquisition function. Many works have adapted Bayesian Optimization for sequence design~\citep{stanton2022accelerating, gruver2024protein, swersky2020amortized, belanger2019biological}. LaMBO~\citep{stanton2022accelerating} uses a MLM style decoding guided with acquisition values to address the multi objective nature of real-world protein design. A follow-up work, LaMBO-2~\citep{gruver2024protein}, argues to use diffusion optimized sampling models in a similar framework for better design. 

\section{Preliminaries}
~\label{prelims}
\paragraph{Reinforcement Learning} In RL, we model a decision making task through an environment that describes a Markov Decision Process (MDP) $<\mathcal{S, A, R, P}>$, where $\mathcal{S}, \mathcal{A}, \mathcal{R}$ and $\mathcal{P}$ represent the state space, action space, rewards and transition dynamics respectively. At every timestep, the agent receives a state $s_t \in \mathcal{S}$ and predicts an action $a_t \in \mathcal{A}$, to transition to a new state according to the transition probability $p(s_{t+1} \mid s_t, a_t)$, and receives a reward $r_t \in \mathcal{R}$. The agent chooses an action using a policy $\pi_{\theta}(a_t|s_t)$, parameterized by $\theta$. The objective is to learn an optimal policy $\pi_{\theta}^*(a_t|s_t)$ that maximises the expected sum of rewards from the environment. Concretely, optimization problem can be summarized as,
\begin{equation*}
    \theta^* = \argmax_{\theta} \mathbb{E}_{a_{0:T} \sim \pi_\theta} \left[ \sum^{T}_{t = 0} \gamma^t r_t \right]
\end{equation*}

\paragraph{Problem Setting}
We frame the problem of protein sequence design in an RL loop, where at each timestep the agent receives a sequence and predicts the mutation to be performed. The sequences are represented through one hot encodings of each character in the sequence. That is, given a sequence of length $L_{s}$ and an alphabet of size $L_{a}$, the state vector is denoted by $s \in \mathbb{R}^{L_{s} \times L_{a}}$. The action space is modelled as a discrete choice over $L_{s} \times L_{a}$ options, indicating the position of the mutation and the mutated character. The mutated sequence is then returned as the next state. The reward is the pTM score of the mutated sequence, which is obtained from either the oracle model or the proxy model. Our environment thus is completely deterministic, the state transitions are simple sequence mutations and the rewards are deterministically obtained from the models. Finally, we design the environment to be an infinite horizon task. RL training is simultaneously conducted on a batch of $100$ sequences. ESMFold is queried with the batch of all $100$ sequences and inference is run across 4 GPUs to obtain the rewards (i.e, pTM score).
% We impose a time limit $T_m$ where $T_m = 2 \times L_{s}$ and reset the sequence to a random sequence after the time limit so that the agent does not overfit to a particular starting sequence.

\begin{figure}
\centering
\begin{subfigure}{.4\textwidth}
  \centering
  \includegraphics[width=\linewidth]{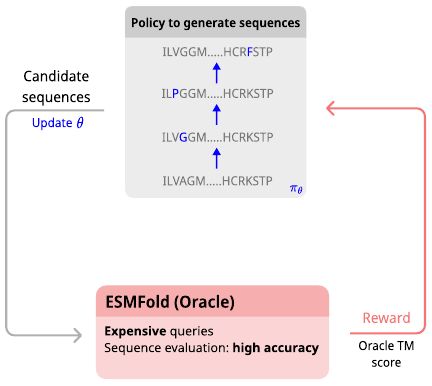}
  \caption{}
  \label{fig:oracle-training-overview}
\end{subfigure}%
\hspace{4em}
\begin{subfigure}{.4\textwidth}
  \centering
  \includegraphics[width=\linewidth]{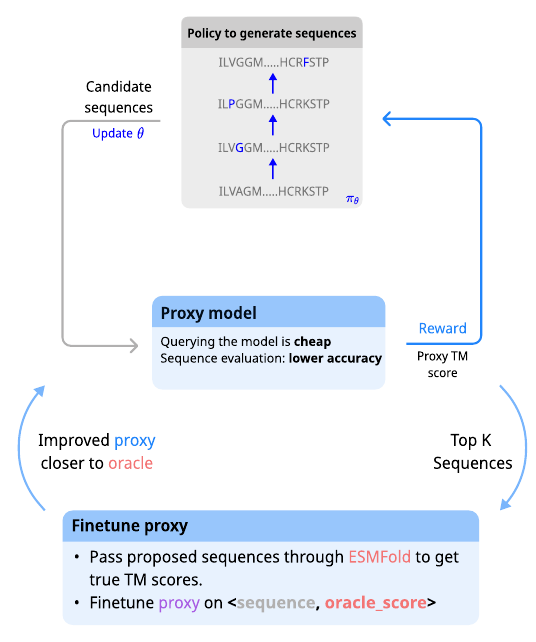}
  \caption{}
  \label{fig:pf-training-overview}
\end{subfigure}
\caption{Training policies with the a) oracle, b) proxy reward models.}
\label{fig:training-overview}
\end{figure}

% \begin{wrapfigure}[20]{R}{7cm} 
%     \centering
%     \includegraphics[width=7cm]{./overview/oracle.pdf}
%     \caption{Training policies with the oracle reward model.}
%     \label{fig:oracle-training-overview}
% \end{wrapfigure}

\section{Reinforcement Learning for Generating Protein Sequences}
\subsection{Approach}~\label{approach}In this paper, we provide a modular approach to leverage existing protein language models for sequence design, in an RL framework. Our work relies on using pretrained LMs and then being able to train a RL based agent for mutating the sequences over long horizon lengths, for generating amino acid sequences. To this end, we first provide an overview of the different RL algorithms being tests for the protein design tasks. We note that even though prior works have explored this, we provide the first approach for RL based algorithms to leverage existing protein language models, such as the well known ESMFold model~\cite{lin2023evolutionary}, for these tasks.

\textbf{Baselines: } We provide an overview of the baselines that were tested on this benchmark. We experiment with both value based and policy based deep RL algorithms for learning mutation policies for protein sequences: (1) Deep Q Network (DQN) \citep{mnih2013playing} (2) Rainbow \citep{hessel2018rainbow} (3) Proximal Policy Optimization (PPO) \citep{schulman2017proximal} (4) Soft Actor-Critic (SAC) \citep{haarnoja2018soft} for the base RL algorithms. We also try an additional exploration based baseline, namley (5) PPO with Random Network Distillation (RND) \citep{burda2018exploration}. We compare our approach with several works that have previously been proposed for similar tasks, namely (6) GFlowNets \citep{bengio2021flow, bengio2023gflownet} and (7) Metropolis Hastings MCMC with simulated annealing ~\citep{metropolis1953equation, hastings1970monte}. Rather than autoregressively constructing the sequences as in~\cite{jain2022biological}, we use a single-step GFlowNet to mutate sequences.

In our setup, starting sequences are randomly initialized and the policy is mutative. At each step, the action sampled from the policy, chooses one of $L_s \times L_a$ actions, describing the index to mutate and also the residue from the alphabet that will substitute the original amino acid. The policy is trained in one of two ways: (1) using the oracle reward model, or (2) using the proxy reward model, which is finetuned every $K=2000$ steps. Figure~\ref{fig:oracle-training-overview} and~\ref{fig:pf-training-overview} illustrate the RL loop with the oracle reward model and proxy reward model with finetuning, respectively. For all methods, we run $100$ environments in parallel.

% \jith{TODO : Write down the algorithm box}
% \riashat{Jith - I think it's quite important to have the algo box!}

% \riashat{I think key for us is to be very clear on how we are using the policies and value functions for these tasks. It may not be obvious to a RL reader, how we adapt a RL based approach with a language model for these tasks. An algorithm box or other detailed outline may help in this section}

\subsection{Learning the Reward Model}

In this section, we outline our approach for learning a proxy to the reward function. As outlined in section~\ref{approach}, the RL based mutation process tries to maximize the reward function at each step. However, it is computationally expensive to evaluate the reward, especially for protein sequences of longer lengths. 

\begin{wrapfigure}[17]{R}{6cm} 
     \centering
     \includegraphics[width=0.3\textwidth]{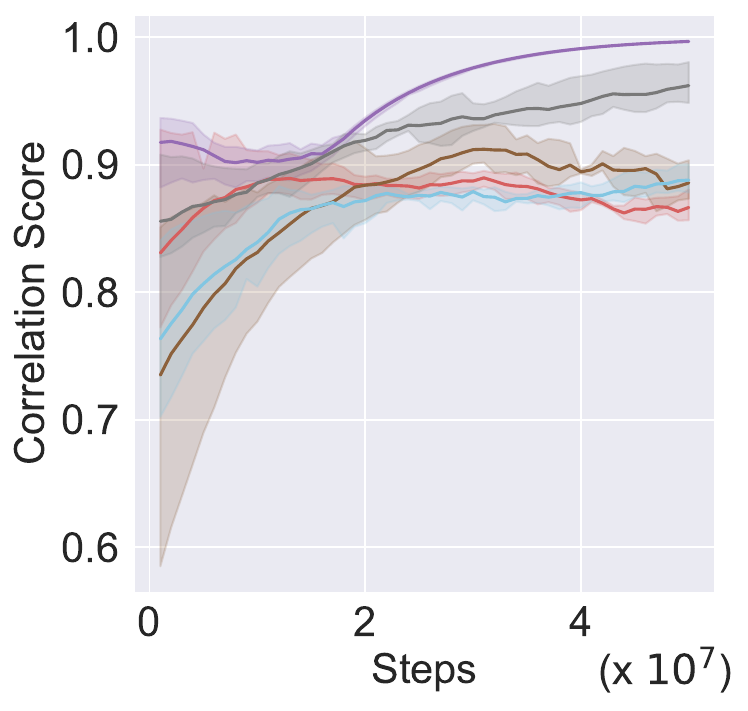}
     \includegraphics[width=0.12\textwidth, trim={0 1cm 0 0},clip]{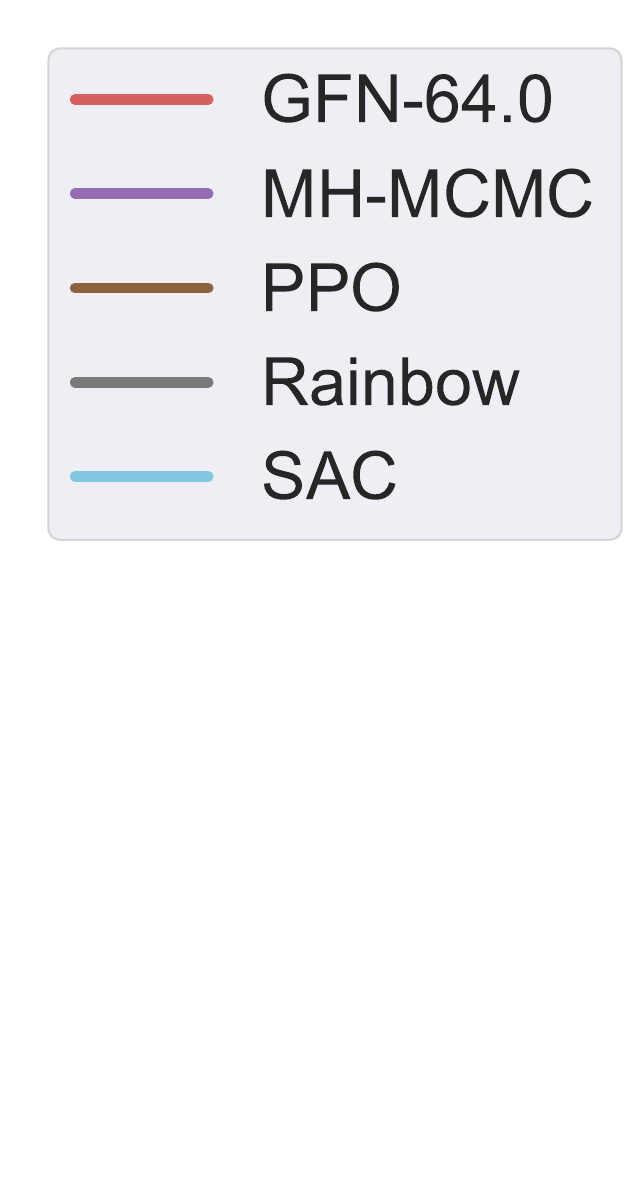}
    \caption{Pearson correlation score between the proxy and oracle model. For readability, the curves are smoothed with exponential moving average.}
    \label{fig:finetune-corr-50}
\end{wrapfigure}

Inspired by knowledge distillation~\citep{hinton2015distilling, polino2018model}, we distill the 3B ESM model (teacher) into a smaller transformer model~\citep{vaswani2017attention}, with $~15 \times 10^3$ parameters, to serve as a proxy reward model. For learning the proxy reward model, we learn the reward function using sequences from the Atlas dataset~\citep{vander2024atlas} since it contains diverse samples. The reward model is then trained using a mean squared regression objective, by predicting the pTM scores for each sequence. 

To ensure we learn a good reward model, we monitor the correlation scores between the proxy predicted scores and the oracle evaluations for each sequence. Additionally, once we have the pretrained proxy model, we can finetune the proxy within the training loop of the algorithms. Every $2000$ environment steps, we compute the proxy pTM scores for the proposed sequences, filter the top $K=100$ sequences, and finetune the proxy on the true pTM scores of these sequences from ESMFold for $50$ epochs. We demonstrate the effectiveness of the proxy finetuning procedure in Figure~\ref{fig:finetune-corr-50}; which shows an increasing trend in Pearson's correlation coefficient between the proxy and oracle model for sequences of length $50$.

% We pretrain the proxy model on sequences in the Atlas dataset \citep{atlas}, performing supervised regression of the pTM scores with the mean squared error. We periodically finetune the proxy model in the RL training loop. Every $2000$ environment steps, we compute the proxy pTM scores for the proposed sequences, filter the top $K=100$ sequences, and finetune the proxy on the true pTM scores of these sequences from ESMFold for $50$ epochs. 

%  In setting (2), the proxy reward model is finetuned on the best $100$ sequences accumulated from the previous $20$ steps (equivalently, from the previous $2000$ sequences).
% We use the ESMFold-3B model as the oracle reward model. For the proxy reward model, we use a simple transformer model with $~15000$ parameters.

% \textit{Fine-Tuning the Reward Model : } The learnt reward model can be used to provide rewards at each step of the mutation process for training the RL policies. However, note that while this is feasible, in practice, it may often be preferred to take a finetuning step where the policy and the reward model are learnt simultaneously \citep{}. In our experiments, we show results with both a pretrained proxy model and a fine-tuning proxy model, where fine-tuning would further update the reward model based on the proposed sequences from the RL algorithms itself. 

\subsection{Evaluating Protein Sequences}
Evaluating the designed protein sequences is inherently multi-objective. In our experiments, evaluations are primarily along two axes, scoring the sequences according to \textbf{biological plausibility} and \textbf{diversity}. Biological plausibility is the useful to assess the viability of the proposed sequences within biological contexts. To this end, we report the mean predicted Template Modeling (pTM) score and the mean predicted Local Distance Difference Test (pLDDT) score across the proposed sequences:

\paragraph{TM score \citep{zhang2004scoring}.} A metric in $(0, 1]$ that evaluates the structural similarity between two proteins, with higher TM-scores indicating better structural similarity. It is a variation of the Levitt-Gerstein (LG) score, which weights shorter distances between residues more strongly than longer distances. The TM-score between a target structure $s_\text{tgt}$ and template structure $s_\text{tmp}$ is defined as:

\begin{align}
    d_{\text{TM}}(s_\text{tgt}, s_\text{tmp}) = \max \bigg[ \frac{1}{L_{\text{tgt}}} \sum\limits_{i=1}^{L} \frac{1}{1 + \big( \frac{d_i}{d_0(L_{\text{tgt}})}\big)^2} \bigg]
\end{align}

where $L_{\text{tgt}}$ is amino acid sequence length of the target protein, $L$ is the number of residues that appear in both template and target structures, $d_i$ is the distance between the $i^{\text{th}}$ residues in template and target structures, and $d_0(L_{\text{tgt}})$ is taken as $1.24 \sqrt[3]{(L_{\text{tgt}} - 15}) - 1.8$ and normalizes sequence length

\paragraph{lDDT score \citep{mariani2013lddt}.} A metric in $(0, 100]$ which provides a per-residue measure of confidence in the local accuracy of a predicted protein structure. lDDT estimates how well the prediction would agree with an experimental structure and a score of above 70 usually corresponds to a correct backbone prediction with misplacement of some side chains. An lDDT score of 90 or greater indicates high confidence in the predicted structure, while a score below 50 indicates poor confidence.

% \paragraph{BLAST } \color{red}{BLAST hits, if any (similar known proteins in sequence space)}

% \paragraph{FoldSeek} \color{red}{FoldSeek hits, if any (similar known proteins in structural space)}

In addition to obtaining high-scoring sequences, we are interested in obtaining \textit{diverse} sequences. Evaluating diversity is multi-dimensional and can be viewed through different lenses such as diversity in sequence space, structural diversity of the protein, number of modes discovered, and hyper-volume. In our experiments, we report multiple metrics for a comprehensive evaluation of sequence diversity through all these lenses:

\paragraph{Diversity in Sequence Space.} The Mean Pairwise Hamming Distance (MP-HD) over a dataset $\mathcal{D}$ of sequences is a standard metric to evaluate the diversity in the sequence space \citep{Angermueller2020Model-based, jain2022biological}, and is measured as:
\begin{align}
     \textbf{MP-HD} = \dfrac{ \sum\limits_{x_i \in \mathcal{D}} \sum\limits_{x_j \in \mathcal{D} \setminus x_i } d_{\text{HD}}(x_i, x_j) }{\lvert \mathcal{D} \rvert \; (\lvert \mathcal{D} \rvert - 1)} 
\end{align}
where $d_{\text{HD}}(., .)$ is the hamming distance between two sequences.

\paragraph{Structural Diversity.} We first obtain the structures $\mathcal{S}$ from $\mathcal{D}$ by performing a forward pass through a protein structure prediction model (ESMFold in our case) and then compute the Mean Pairwise TM score (MP-TM) and Mean Pairwise RMSD (MP-RMSD). 

\begin{minipage}{0.45\linewidth}
    \begin{equation}
        \textbf{MP-TM} = \dfrac{\sum\limits_{s_i \in \mathcal{S}} \sum\limits_{s_j \in \mathcal{S} \setminus s_i } {d_\text{TM}}(s_i, s_j) }{\lvert \mathcal{D} \rvert \; (\lvert \mathcal{D} \rvert - 1)}
    \end{equation}
\end{minipage}%
\hfill
\begin{minipage}{0.5\linewidth}
    \begin{equation}
        \textbf{MP-RMSD} = \dfrac{\sum\limits_{s_i \in \mathcal{S}} \sum\limits_{s_j \in \mathcal{S} \setminus s_i } d_{\text{RMSD}}(s_i, s_j) }{\lvert \mathcal{D} \rvert \; (\lvert \mathcal{D} \rvert - 1)}
    \end{equation}
\end{minipage}

Since TM-score is a closeness metric and RMSD is a distance metric, we typically want \textbf{low MP-TM} ($\downarrow$) and \textbf{high MP-RMSD} ($\uparrow$).

However, the reward models could be imperfect or misspecified. To quantify this, we compare outputs of ESMFold with that of AlphaFold. Concretely, we compare the mean pTM and pLDDT scores predicted by each model across the proposed sequences for each algorithm. We also report the mean RMSD between the structure predicted by both models, over all proposed sequences.

% \todoexp{Follow this up with the experiment; what are the scores of the proposed sequences under AlphaFold for example. Make a table similar to ones we have.}

% \riashat{Discuss DockQ scores here : correlation plots between ESMFold and AlphaFold when folding the proposed sequences}

% {\color{red}{TODO: Overkill but would be cool to see correlation plots between ESMFold and AlphaFold, for PTM plDDT on the proposed sequences for each algo? Can be added in Appendix}}

\section{Experiments and Analysis}
In this section, we outline our experiments for sequence design. In all our experiments, we use ESMFold either directly as the reward model or as a signal to finetune the proxy model. Our main goal is to test the efficacy of different sequence design algorithms. Through our experiments, we analyze the performance of all methods and ask the following questions.

\begin{wrapfigure}[19]{R}{7cm}
    \centering
    \includegraphics[width=0.32\textwidth]{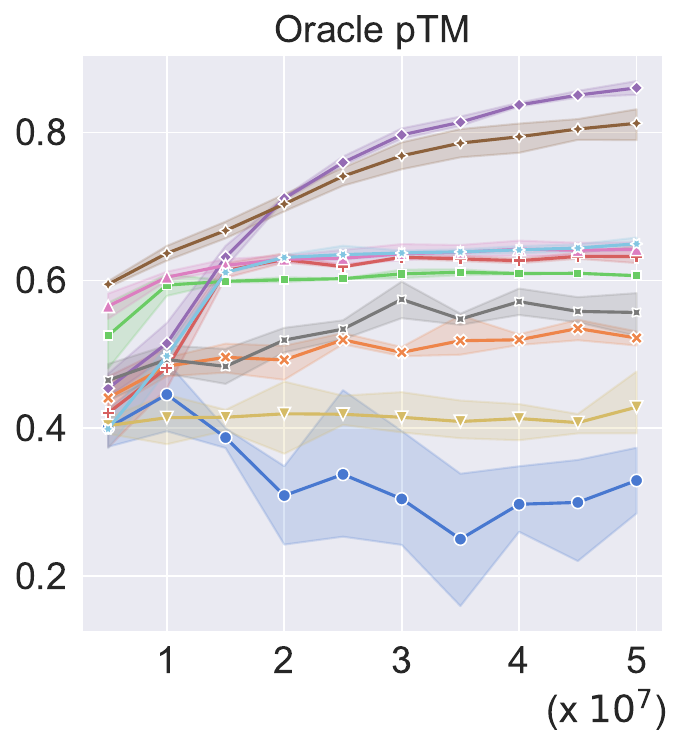}
    \includegraphics[width=0.15\textwidth, trim={0 17.2cm 0 0},clip]{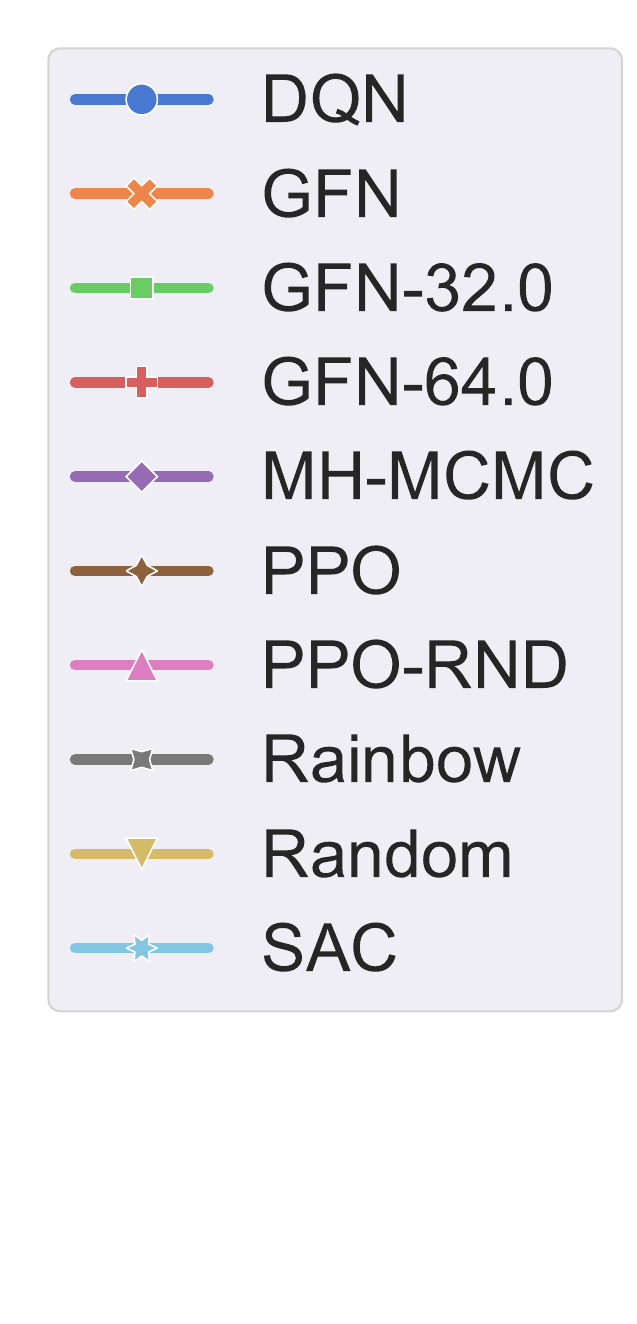}
    \caption{Learning curves for optimizing the reward (pTM) from ESMFold on sequences of length 50. The x-axis is the number of queries to the reward model.}
    \label{fig:50seqlen-oracle-inf-traincurves}
\end{wrapfigure}

\textbf{\emph{How effectively do various sequence design algorithms optimize the reward from ESMFold?}}

Table~\ref{tab:oracle-inf-50} and~\ref{tab:oracle-inf-100} (appendix) present the results for all methods evaluated in the infinite horizon setting for sequences of lengths $50$ and $100$, respectively. The reward used in these experiments is the pTM scores obtained from the oracle model (ESMFold). All results are averaged over three random seeds, and all experiments for this analysis were run on 4 A100 GPUs. The results in these tables highlight the effectiveness of RL algorithms and GFlowNets for sequence design. Figure~\ref{fig:50seqlen-oracle-inf-traincurves} illustrates the training curves for optimizing the oracle reward for sequences of length $50$.

Our training results show MCMC outperforms PPO in terms of the pTM score. While this result might suggest that MCMC is superior for sequence design, it is important to consider the broader context and practical implications. 

\input{./tables/oracle-inf-50-table}

In contrast, learning a policy, such as with RL or GFlowNets, offers several compelling advantages:

\begin{enumerate}
    \item \textbf{Generalization}: Once a policy is trained, it can generalize to new unseen sequences without requiring retraining from scratch. This ability to adapt to new sequences without starting over is crucial for efficiency.
    \item \textbf{Cost efficient at evaluation time}: After training, the policy can generate multiple candidates simply by sampling actions from the policy. This process does not require access to the reward model and incurs only a fraction of the original training cost, making it much more efficient for large-scale sequence generation.
    \item \textbf{Adaptability}: The policy can be finetuned with additional data, allowing for iterative improvements and adaptations to new requirements or environments. This flexibility is not easily achievable with MCMC.
\end{enumerate}

Although MCMC shows better performance in terms of the pTM score, it has significant limitations. Specifically, MCMC lacks the ability to generalize to unseen sequences and must be run from scratch for each new sequence. This leads to increased computational costs and time for every new instance. 

\textbf{\emph{Can a smaller, cost-efficient proxy model be trained to approximate ESMFold scores accurately? Can sequences be generated by optimizing the rewards from this proxy model?}}

We find that all RL methods and GFlowNets are able to maintain high oracle pTM scores by optimizing on the proxy pTM (Fig ~\ref{fig:50seqlen-proxy-inf-traincurves} and Table~\ref{tab:pf-inf-50}), while \textbf{cutting down the runtime by an order of magnitude} (Table ~\ref{tab:runtimes-inf-50}). All results are averaged over three random seeds, and all experiments for this analysis were run on a single RTX8000 GPU. We present the complete results in Table~\ref{tab:pf-inf-50}.

Most notably, while MCMC demonstrated the best performance when optimizing directly on the oracle (Tab.~\ref{tab:oracle-inf-50}), it tends to overcommit to the proxy scores, ultimately compromising its performance on the oracle pTM (Fig ~\ref{fig:50seqlen-proxy-inf-traincurves}). We hypothesize that MCMC, being a hill-climbing method, may become trapped in local optima that are favorable according to the proxy model but not according to the oracle. Given that the proxy is a smaller, less complex model, its reward landscape likely contains different local optima that do not align with those of the oracle.

\begin{minipage}[h]{0.65\textwidth}
    \centering
    \includegraphics[width=0.49\textwidth]{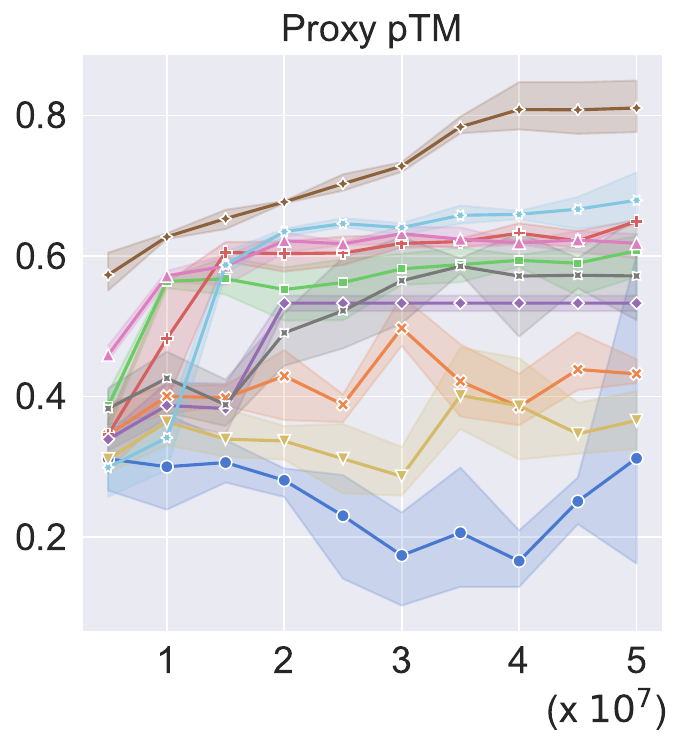}
    \includegraphics[width=0.49\textwidth]{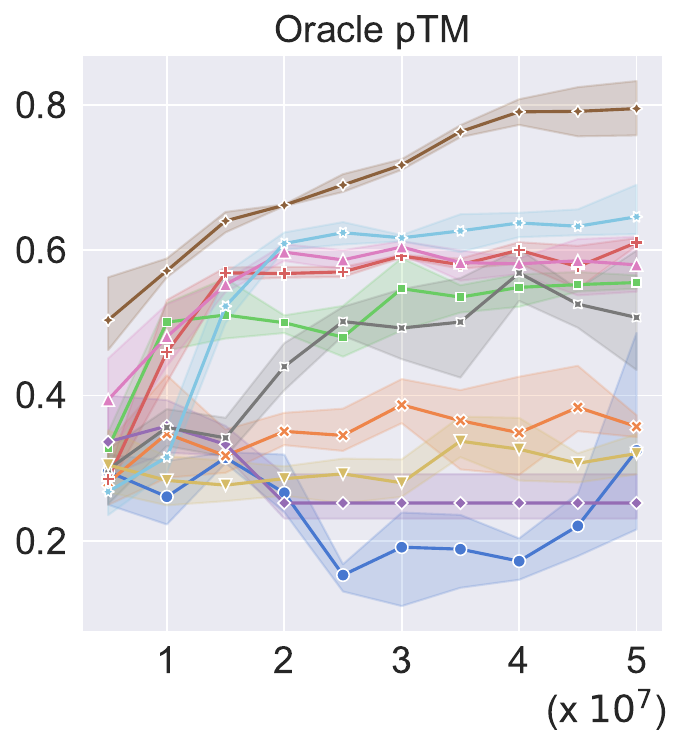}
    \includegraphics[width=0.95\textwidth]{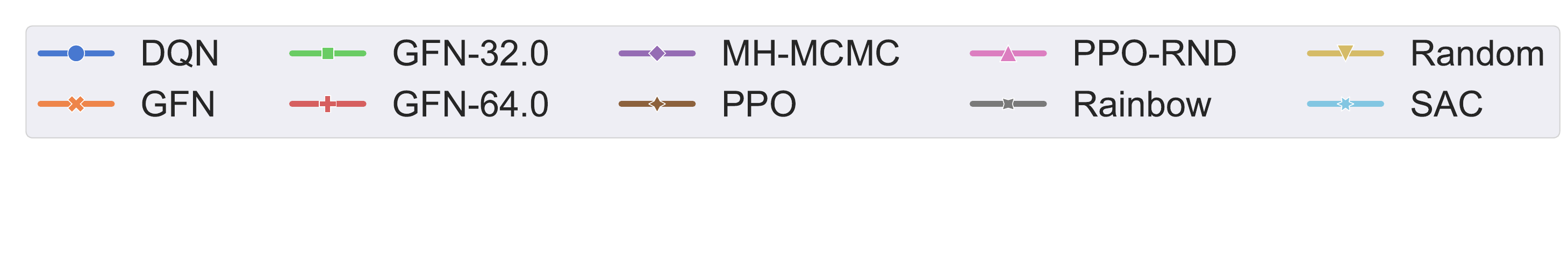}
    \captionof{figure}{Learning curves for optimizing the reward from the proxy model on sequence length $50$. Both the proxy pTM and oracle pTM are shown, with the x-axis referring to number of proxy reward queries.}
    \label{fig:50seqlen-proxy-inf-traincurves}
\end{minipage}\hspace{0.04\textwidth}
\begin{minipage}[t]{0.3\textwidth}
    \centering
    \vspace{-9em}
    \captionof{table}{Comparing A100 GPU runtimes, in minutes, for oracle and proxy-finetune experiments (Sequence length $50$). Proxy runtimes are faster by \emph{an order of magnitude}.}
    \begin{adjustbox}{center,scale=0.9}
    \begin{tabular}{lcc}
        \toprule
        & \textbf{\color{red!65}{\shortstack{Oracle \\ Runtime}}}
        & \textbf{\color{blue!65}{\shortstack{Proxy \\ Runtime}}} \\
        \cmidrule(lr){2-3}
        PPO & $1200$ & $75$ \\
        PPO-RND & $1270$ & $80$ \\
        DQN & $2400$ & $345$ \\
        SAC & $2640$ & $285$ \\
        Rainbow & $2760$ & $300$ \\
        GFN & $988$ & $75$ \\
        \bottomrule
    \end{tabular}
    \end{adjustbox}
    \label{tab:runtimes-inf-50}
\end{minipage}

% \shiva{DQN performs worse than random. Check if DQN with prioritized experience replay performs better?}

\input{./tables/pf-inf-50-table}

\textbf{\emph{How do different algorithms compare when holistically evaluated on biological plausibility and diversity scores?}}

In both experiments, whether optimizing on the oracle or the proxy reward model, the generated sequences are evaluated using five metrics: pTM and pLDDT scores for biological plausibility, two structural diversity metrics (MP-TM and MP-RMSD), and MP-HD, which measures diversity in the sequence space. Table~\ref{tab:oracle-inf-50} and~\ref{tab:pf-inf-50} summarizes all five metrics for both settings. 

\begin{figure}[t]
     \centering
     \includegraphics[width=0.325\textwidth]{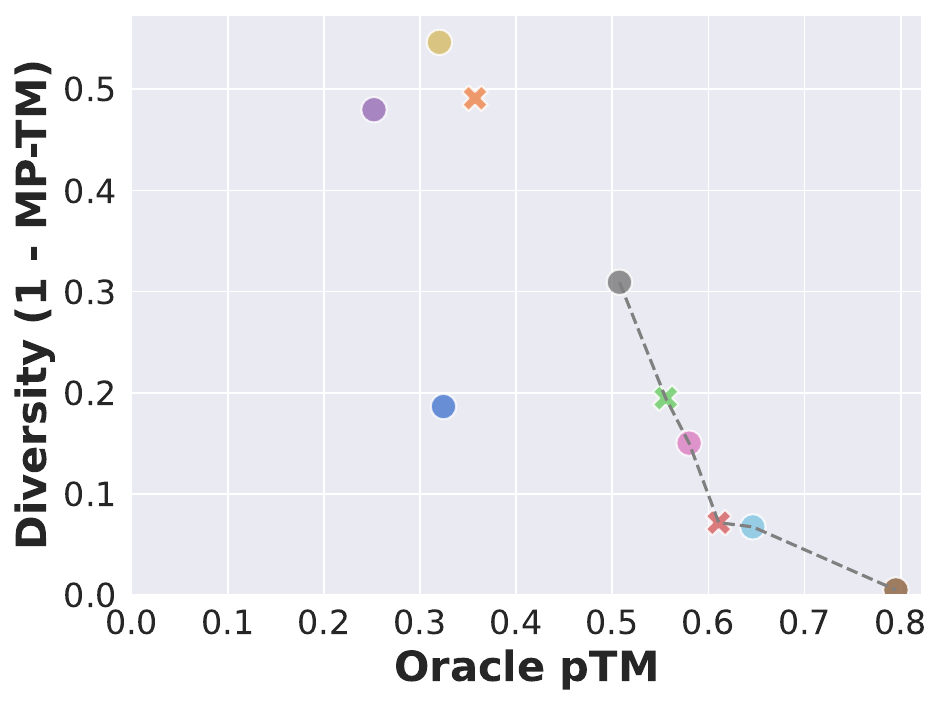}
     \includegraphics[width=0.325\textwidth]{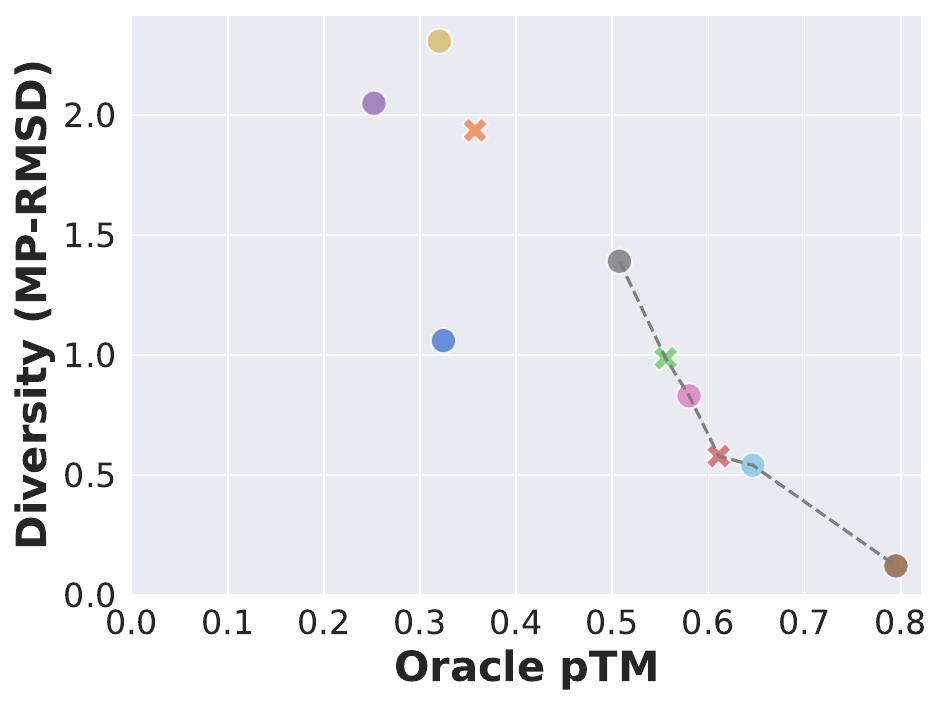}
     \includegraphics[width=0.325\textwidth]{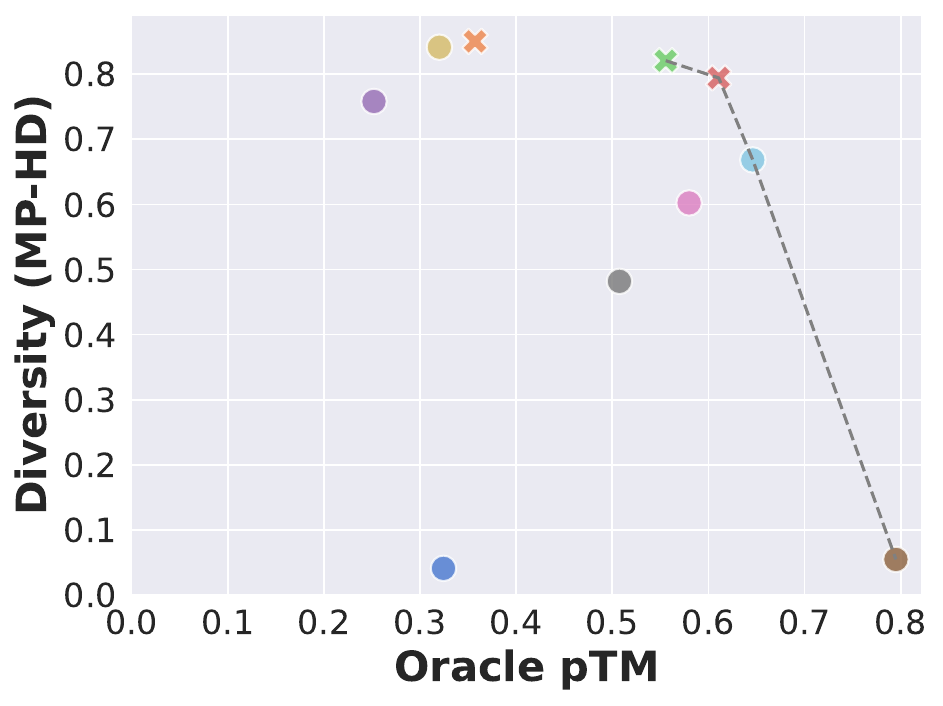}
     \includegraphics[width=0.95\textwidth]{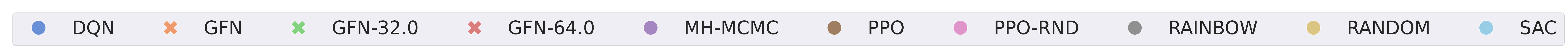}
    \caption{\textbf{Pareto plots for optimization on the finetuned {\color{blue!70}{proxy}} scores:} Plotted across two structural diversity metrics and one sequence diversity metric (MP-HD). Only methods with atleast $0.5$ pTM are highlighted. \textbf{PPO}, \textbf{SAC}, and \textbf{GFlowNets} form the Pareto front across all 3 diversity metrics.}
    \label{fig:pareto-50-pf-inf}
\end{figure}

Since the evaluation of the generated sequences is inherently multi-objective, we visualize Pareto plots, combining one biological plausibility metric (in this case, the pTM score) with each of the three diversity metrics. Figures~\ref{fig:pareto-50-pf-inf} and~\ref{fig:pareto-50-oracle-inf} provide these visualizations, highlighting only those methods that achieve a pTM score of at least $0.5$.

From Figure~\ref{fig:pareto-50-pf-inf} it is evident that PPO, SAC, and GFN ($\beta=32, 64$) are the top-performing methods, establishing the Pareto front across all three diversity metrics. Notably, Rainbow also appears on the Pareto front for structural diversity metrics such as MP-TM and MP-RMSD, although it does not achieve this for MP-HD. Additionally, GFlowNets exhibit flexibility in balancing pTM scores with diversity. By adjusting the hyperparameter $\beta$, GFlowNets can attain different positions on the Pareto front, as highlighted by the points marked with an "x" in Fig~\ref{fig:pareto-50-pf-inf}).

While MCMC remains in the Pareto set for sequences of length 50, demonstrating good diversity and bio-plausibility scores, we observe a decline in its diversity as we scale to longer sequences. For sequences of length 100, PPO becomes comparable to MCMC in terms of both bio-plausibility scores and diversity (Table~\ref{tab:oracle-inf-100} on sequence length $100$).

\emph{\textbf{How does the setting (infinite horizon vs finite horizon) affect performance?}}

\begin{wrapfigure}[20]{R}{9cm} 
     \centering
     \includegraphics[width=0.3\textwidth]{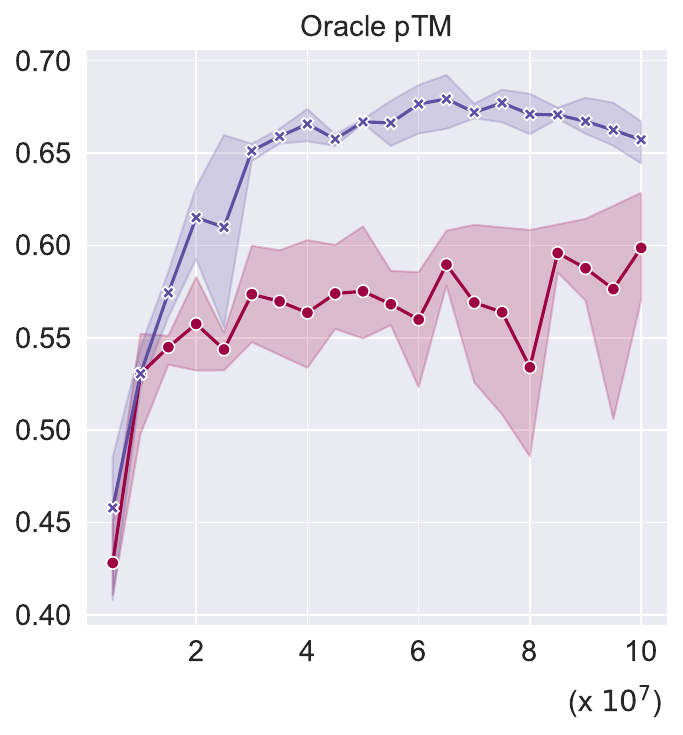}
     \includegraphics[width=0.3\textwidth]{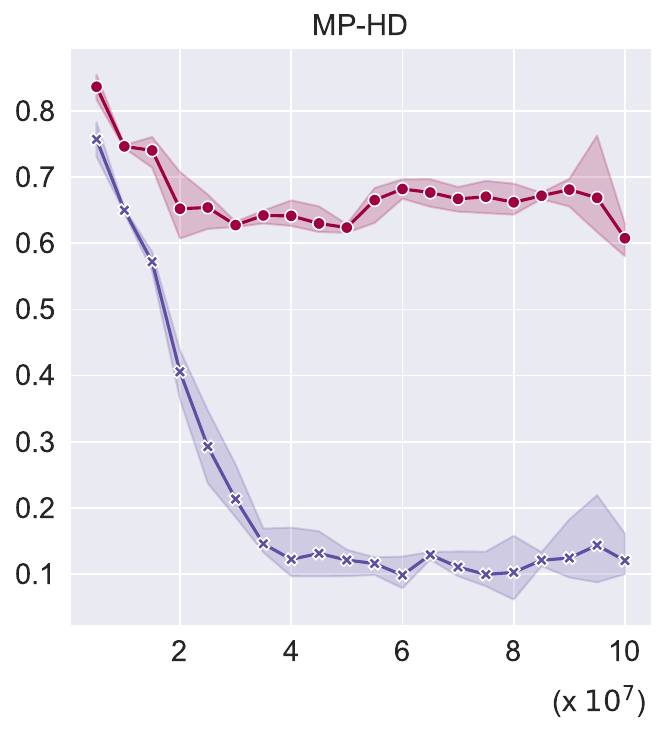}
     \hfill
     \includegraphics[width=0.2\textwidth]{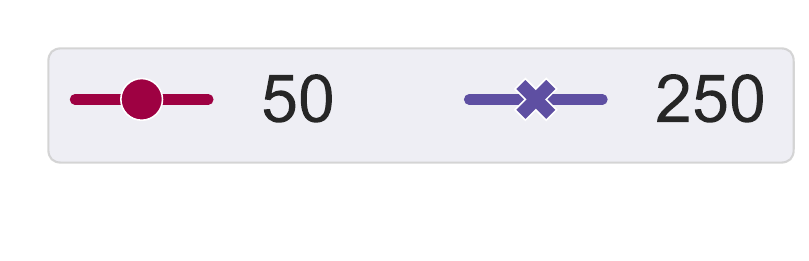}
    \vspace{-0.3cm}
    \caption{Effect of episode length in the finite horizon setting on the pTM-diversity tradeoff. The ablation is performed for PPO trained on the ESM-PF proxy model for sequences of length $50$.}
    \label{fig:episode-len-ppo-ablation}
\end{wrapfigure}

Here, we perform an ablation study on episode length in the finite horizon setting, contrasting it with our previous results in the infinite horizon setting. Due to brevity and computational limitations, we conduct experiments solely on PPO optimized with ESM-PF on sequences of length $50$. We examine different episode lengths (i.e., $L_s$ and $5L_s$) and observe their effects on the oracle pTM score and diversity metric. For conciseness, we consider only the MP-HD for diversity. Generally, for the same reward budget, shorter episode lengths result in low pTM and high diversity, whereas longer episode lengths result in high pTM and low diversity (Figure~\ref{fig:episode-len-ppo-ablation}). Depending on the confidence in the reward model, the RL loop can be run in different settings to trade off diversity (accounting for reward misspecification) for bio-plausibility scores, and vice-versa.

\section{Limitations}
Our results are constrained by sequence length due to the prohibitive computational requirements associated with longer sequences. The findings presented in this paper rely on either the proxy or the 3B ESMFold oracle model. However, the reward model may itself be uncertain or misspecified. Future research should consider extending these results to optimization on other PLMs, such as AlphaFold2, or the larger 15B parameter ESMFold model. Additionally, exploring larger proxy reward models could provide a more accurate approximation of the oracle reward model for longer sequences. Though we do not foresee any negative consequences arising from this work, PLMs could be used maliciously by users.

\section{Conclusion}
In this work, we propose to leverage existing pretrained PLMs to learn a mutation policy for generating protein sequences and demonstrate the efficacy of this approach. We benchmark the performance of the algorithms in a wide range of settings -- training with the $3$B ESMFold and the proxy, generating various sequence lengths, and studying the pTM-diversity tradeoff for different episode lengths. To this end, we demonstrate and compare performance, of several existing deep RL algorithms compared to well known baselines, and show that RL can be quite effective in this domain. We share a modular implementation in the appendix, with support for adding a PLM of the user's choice as the reward model, as well as adding other RL algorithms.

\section*{Acknowledgements}

The authors would like to thank Compute Canada and Mila for the computing resources required for this work. JS would like to thank Kausthubh Ramachandran, Michael Jendrusch, Moksh Jain and Nanda Harishankar Krishna for fruitful discussions and for providing feedback on initial drafts of the paper. DN and SEK would also like to acknowledge CIFAR.

% \paragraph{Future Work} \riashat{Rewrite} Several works have explored sequence design through the lens of diffusion or flow matching based approaches, compared to mutation based approach with RL or GFNs. In future, it would be useful to experimentally compare these approaches, to test effectiveness of RL based algorithms compared to discrete generative approaches. 

\bibliographystyle{unsrtnat}
\bibliography{neurips_2024}

\newpage
\section*{NeurIPS Paper Checklist}

\begin{enumerate}

\item {\bf Claims}
    \item[] Question: Do the main claims made in the abstract and introduction accurately reflect the paper's contributions and scope?
    \item[] Answer: \answerYes{} % Replace by \answerYes{}, \answerNo{}, or \answerNA{}.
    \item[] Justification: We show how PLMs can be leveraged for protein sequence design in an RL paradigm and provide comprehensive evaluation of RL baselines across both sequence-based and structural metrics. %\justificationTODO{}

\item {\bf Limitations}
    \item[] Question: Does the paper discuss the limitations of the work performed by the authors?
    \item[] Answer: \answerYes{} % Replace by \answerYes{}, \answerNo{}, or \answerNA{}.
    \item[] Justification: We provide a section on the limitations of the current work and potential extentions to address these limitations.

\item {\bf Theory Assumptions and Proofs}
    \item[] Question: For each theoretical result, does the paper provide the full set of assumptions and a complete (and correct) proof?
    \item[] Answer: \answerNA{} % Replace by \answerYes{}, \answerNo{}, or \answerNA{}.
    \item[] Justification: The paper does not include theoretical results. % \justificationTODO{}

    \item {\bf Experimental Result Reproducibility}
    \item[] Question: Does the paper fully disclose all the information needed to reproduce the main experimental results of the paper to the extent that it affects the main claims and/or conclusions of the paper (regardless of whether the code and data are provided or not)?
    \item[] Answer: \answerYes{} % Replace by \answerYes{}, \answerNo{}, or \answerNA{}.
    \item[] Justification: We provide the code for the RL baselines and environment implementations used in the paper. The repositories include instructions on how to reproduce all the results in the paper. %\justificationTODO{}

\item {\bf Open access to data and code}
    \item[] Question: Does the paper provide open access to the data and code, with sufficient instructions to faithfully reproduce the main experimental results, as described in supplemental material?
    \item[] Answer: \answerYes{} % Replace by \answerYes{}, \answerNo{}, or \answerNA{}.
    \item[] Justification: The code to reproduce the experiments has been added to the supplementary material. % \justificationTODO{}

\item {\bf Experimental Setting/Details}
    \item[] Question: Does the paper specify all the training and test details (e.g., data splits, hyperparameters, how they were chosen, type of optimizer, etc.) necessary to understand the results?
    \item[] Answer: \answerYes{} % Replace by \answerYes{}, \answerNo{}, or \answerNA{}.
    \item[] Justification: The experimental section contains the problem setting and a description of how the code can be used. % \justificationTODO{}

\item {\bf Experiment Statistical Significance}
    \item[] Question: Does the paper report error bars suitably and correctly defined or other appropriate information about the statistical significance of the experiments?
    \item[] Answer: \answerYes{} % Replace by \answerYes{}, \answerNo{}, or \answerNA{}.
    \item[] Justification: We run all experiments for multiple seeds and report the mean and standard deviation across runs. % \justificationTODO{}

\item {\bf Experiments Compute Resources}
    \item[] Question: For each experiment, does the paper provide sufficient information on the computer resources (type of compute workers, memory, time of execution) needed to reproduce the experiments?
    \item[] Answer: \answerYes{} % Replace by \answerYes{}, \answerNo{}, or \answerNA{}.
    \item[] Justification: We include the number of GPUs used per experiment and the average running time. % \justificationTODO{}
    
\item {\bf Code Of Ethics}
    \item[] Question: Does the research conducted in the paper conform, in every respect, with the NeurIPS Code of Ethics \url{https://neurips.cc/public/EthicsGuidelines}?
    \item[] Answer: \answerYes{} % Replace by \answerYes{}, \answerNo{}, or \answerNA{}.
    \item[] Justification: Yes, all authors have reviewed the Code of Ethics. % \justificationTODO{}

\item {\bf Broader Impacts}
    \item[] Question: Does the paper discuss both potential positive societal impacts and negative societal impacts of the work performed?
    \item[] Answer: \answerYes{} % Replace by \answerYes{}, \answerNo{}, or \answerNA{}.
    \item[] Justification: We address potential harms in the limitations section. 
    
\item {\bf Safeguards}
    \item[] Question: Does the paper describe safeguards that have been put in place for responsible release of data or models that have a high risk for misuse (e.g., pretrained language models, image generators, or scraped datasets)?
    \item[] Answer: \answerNA{} % Replace by \answerYes{}, \answerNo{}, or \answerNA{}.
    \item[] Justification: The work in this paper poses no such risks.

\item {\bf Licenses for existing assets}
    \item[] Question: Are the creators or original owners of assets (e.g., code, data, models), used in the paper, properly credited and are the license and terms of use explicitly mentioned and properly respected?
    \item[] Answer: \answerYes{} % Replace by \answerYes{}, \answerNo{}, or \answerNA{}.
    \item[] Justification: We cite the papers that introduce the PLMs used in this paper. % \justificationTODO{}

\item {\bf New Assets}
    \item[] Question: Are new assets introduced in the paper well documented and is the documentation provided alongside the assets?
    \item[] Answer: \answerYes{} % Replace by \answerYes{}, \answerNo{}, or \answerNA{}.
    \item[] Justification: We document the code released in a README. % \justificationTODO{}

\item {\bf Crowdsourcing and Research with Human Subjects}
    \item[] Question: For crowdsourcing experiments and research with human subjects, does the paper include the full text of instructions given to participants and screenshots, if applicable, as well as details about compensation (if any)? 
    \item[] Answer: \answerNA{} % Replace by \answerYes{}, \answerNo{}, or \answerNA{}.
    \item[] Justification: The paper does not involve crowdsourcing nor research with human subjects. %\justificationTODO{}

\item {\bf Institutional Review Board (IRB) Approvals or Equivalent for Research with Human Subjects}
    \item[] Question: Does the paper describe potential risks incurred by study participants, whether such risks were disclosed to the subjects, and whether Institutional Review Board (IRB) approvals (or an equivalent approval/review based on the requirements of your country or institution) were obtained?
    \item[] Answer: \answerNA{} % Replace by \answerYes{}, \answerNo{}, or \answerNA{}.
    \item[] Justification: The paper does not involve crowdsourcing nor research with human subjects. % \justificationTODO{}

\end{enumerate}

%%%%%%%%%%%%%%%%%%%%%%%%%%%%%%%%%%%%%%%%%%%%%%%%%%%%%%%%%%%%

\newpage
\appendix

\section{Additional Figures}

\begin{figure*}[!h]
     \centering
     \includegraphics[width=0.325\textwidth]{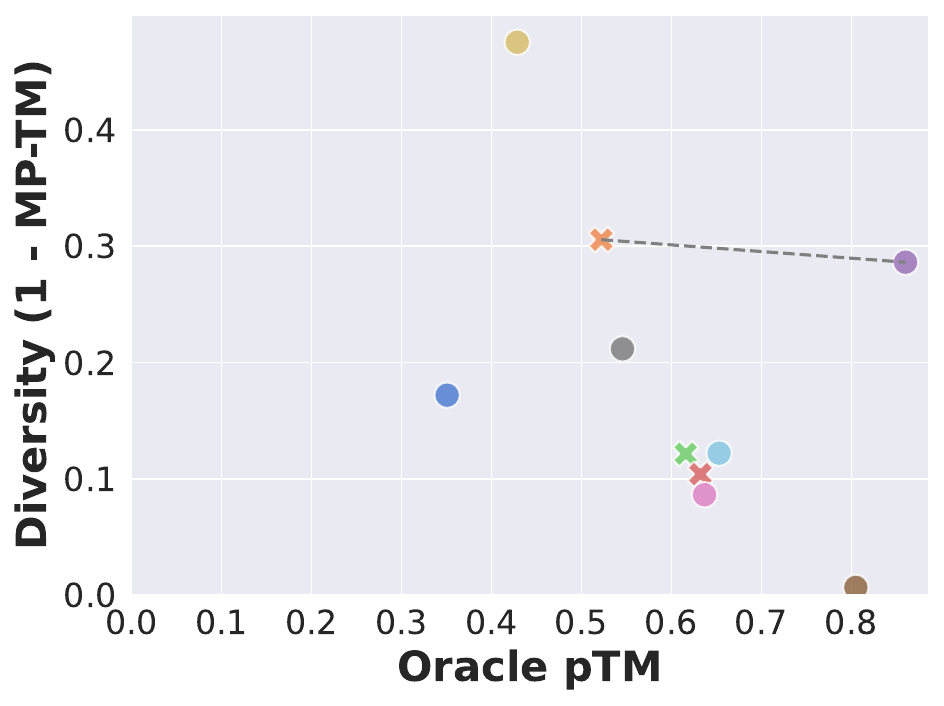}
     \includegraphics[width=0.325\textwidth]{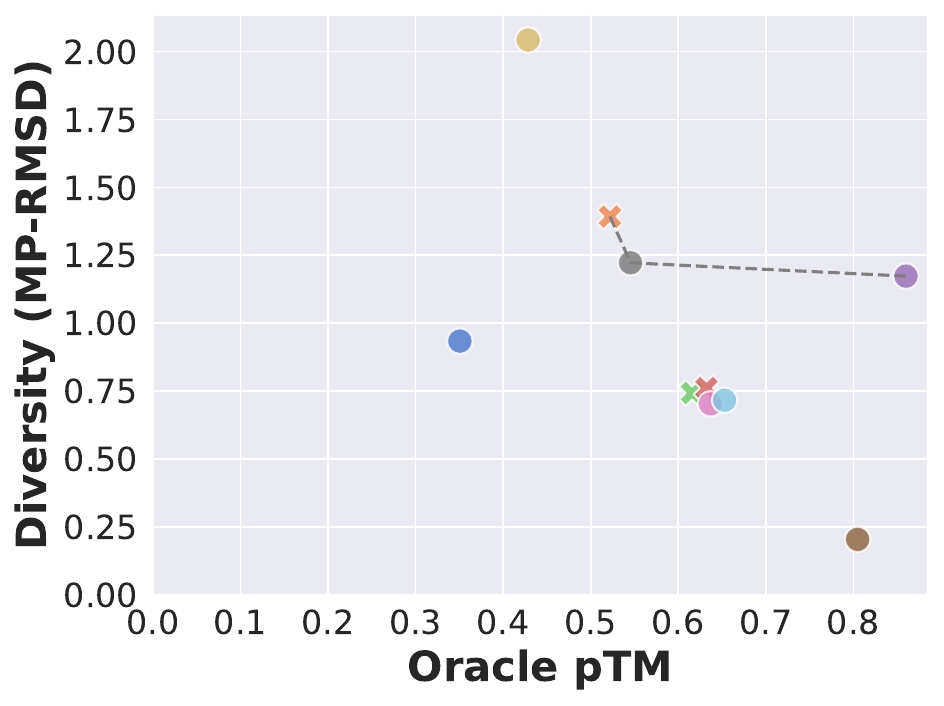}
     \includegraphics[width=0.325\textwidth]{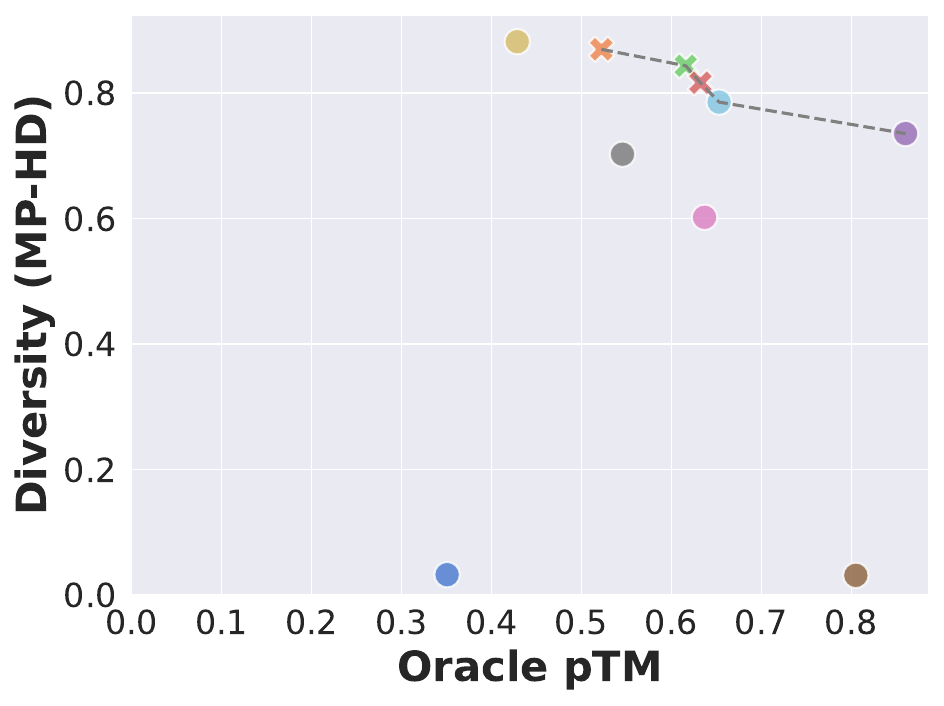}
     \includegraphics[width=0.95\textwidth]{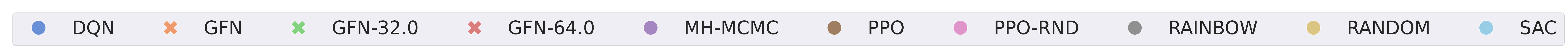}
    \caption{\textbf{Pareto plots for optimizing proteins of length $50$ on {\color{red!70}{ESMFold}} pTM scores:} Plotted across two structural diversity metrics and one sequence diversity metric (MP-HD). Only methods with atleast $0.5$ pTM are highlighted.}
    \label{fig:pareto-50-oracle-inf}
\end{figure*}

\section{Experiments on Longer Sequence Lengths}
\input{./tables/oracle-inf-100-table}
\input{./tables/oracle-inf-150-table}

\section{Biophysical Measures}
In this section, we present several biophysical properties computed from the amino acid sequences of the generated proteins using Biopython~\citep{Cock2009}. Namely, we report the \textbf{molecular weight} ($W_{mol}$), \textbf{instability index}, \textbf{isoelectric point}, \textbf{grand average of hydropathy (GRAVY)} computed with the Kyte-Doolittle scale~\cite{Kyte1982}. These properties are then compared to a reference distribution defined by naturally occurring proteins from the Protein Data Bank (PDB)~\citep{Burley2018} using the Distributional Conformity Score (DCS) introduced in~\cite{frey2023protein}. Tables~\ref{tab:oracle-inf-50-biometrics}, \ref{tab:oracle-inf-100-biometrics}, and \ref{tab:oracle-inf-150-biometrics} report the metrics for proteins of residue lengths $50, 100$ and $150$ respectively.

\input{tables/oracle-inf-50-biometrics}
\input{tables/oracle-inf-100-biometrics}
\input{tables/oracle-inf-150-biometrics}

\input{./tables/pf-dcs}

% \section{Amino Acid Distribution}
\input{./tables/oracle-aa_dist-mae}
\input{./tables/pf-50-mae}

%%%%%%%%%%%%%%%%%%%%%%%%%%%%%%%%%%%%%%%%%%%%%%%%%%%%%%%%%%%%

\end{document}

%% file: tables/oracle-inf-50-table.tex
\setlength\tabcolsep{5pt}
\begin{table}[t]
    \centering
    \caption{\textbf{Optimizing on pTM scores from {\color{red!70}{ESMFold}} (Sequence length $50$)}: Final scores of the generated protein sequences with a budget of $5 \times 10^7$ oracle reward evaluations. \emph{Best} protein confidence scores and \emph{second best} diversity scores are highlighted. All scores are computed on the outputs of the 3B ESMFold model. }
    \begin{adjustbox}{center,scale=0.9}
    \begin{tabular}{lccccccccccc}
        \toprule
         & & \multicolumn{3}{c}{\textbf{Protein Confidence}} 
         & & \multicolumn{5}{c}{\textbf{Diversity Score}} \\
        \cmidrule(lr){3-5}
        \cmidrule(lr){7-11}
         & & pTM ($\uparrow$) 
         & & pLDDT ($\uparrow$) 
         & & MP-TM ($\downarrow$) 
         & & MP-RMSD ($\uparrow$) 
         & & MP-HD ($\uparrow$)  \\
        \midrule

        Random 
        & & \g{0.43}{0.04} 
        & & \g{63.0}{6.21}
        & & \g{0.52}{0.08}
        & & \g{2.04}{0.35} 
        & & \g{0.88}{0.01}
        \\

        MH-MCMC 
        & & \oraclehighlight{\g{0.86}{0.01}}
        & & \oraclehighlight{\g{92.9}{0.17}}
        & & \g{0.71}{0.15} 
        & & \g{1.17}{0.52} 
        & & \g{0.74}{0.04} 
        \\

        DQN 
        & & \g{0.35}{0.04}
        & & \g{56.14}{3.53}
        & & \g{0.83}{0.15}
        & & \g{0.93}{0.69} 
        & & \g{0.03}{0.03}
        \\

        SAC
        & & \g{0.65}{0.01}
        & & \g{85.1}{0.47}
        & & \g{0.88}{0.05}
        & & \g{0.72}{0.08} 
        & & \g{0.79}{0.01}
        \\

        PPO
        & & {\g{0.81}{0.02}}
        & & {\g{91.4}{2.05}}
        & & \g{0.99}{0.01}
        & & \g{0.21}{0.33} 
        & & \g{0.03}{0.05}
        \\

        PPO-RND 
        & & \g{0.64}{0.01}
        & & \g{84.43}{0.75}
        & & \g{0.91}{0.05}
        & & \g{0.70}{0.09} 
        & & \g{0.60}{0.04}
        \\

        Rainbow
        & & \g{0.55}{0.03}
        & & \g{77.7}{3.56}
        & & \g{0.79}{0.06}
        & & \g{1.22}{0.17} 
        & & \g{0.70}{0.02}
        \\

        GFN $(\beta = 6)$ 
        & & \g{0.52}{0.01}
        & & \g{74.0}{1.00}
        & & \oraclehighlight{\g{0.69}{0.03}}
        & & \oraclehighlight{\g{1.39}{0.05}}
        & & \oraclehighlight{\g{0.87}{0.01}}
        \\

        GFN $(\beta = 32)$ 
        & & \g{0.62}{0.02}
        & & \g{82.5}{0.73}
        & & \g{0.88}{0.02}
        & & \g{0.74}{0.09}
        & & \g{0.84}{0.02}
        \\

        GFN $(\beta = 64)$
        & & \g{0.63}{0.01}
        & & \g{83.7}{0.55}
        & & \g{0.90}{0.02}
        & & \g{0.76}{0.04}
        & & \g{0.82}{0.01}
        \\
        
        \bottomrule
    \end{tabular}
    \end{adjustbox}
    \label{tab:oracle-inf-50}
\end{table}

%% file: tables/pf-inf-50-table.tex
\setlength\tabcolsep{5pt}
\begin{table}[h]
    \centering
    \caption{\textbf{Optimizing on pTM scores from the finetuned {\color{blue!70}{proxy}} model (ESM-PF; Sequence length $50$)}: Final scores of the generated protein sequences with a budget of $5 \times 10^7$ proxy reward evaluations. \emph{Best} protein confidence scores and \emph{second best} diversity scores are highlighted. All scores are computed on the outputs of the 3B ESMFold model.}
    \begin{adjustbox}{center,scale=0.9}
    \begin{tabular}{lccccccccccc}
        \toprule
         & & \multicolumn{3}{c}{\textbf{Protein Confidence}} 
         & & \multicolumn{5}{c}{\textbf{Diversity Score}} \\
        \cmidrule(lr){3-5}
        \cmidrule(lr){7-11}
         & & pTM ($\uparrow$) 
         & & pLDDT ($\uparrow$) 
         & & MP-TM ($\downarrow$) 
         & & MP-RMSD ($\uparrow$) 
         & & MP-HD ($\uparrow$)  \\
        \midrule
        
        Random
        & & \g{0.32}{0.03} 
        & & \g{54.0}{2.9}
        & & \g{0.45}{0.02}
        & & \g{2.31}{0.12} 
        & & \g{0.84}{0.00} 
        \\
        
        MH-MCMC 
        & & \g{0.25}{0.03} 
        & & \g{48.6}{1.8}
        & & \g{0.52}{0.06}
        & & \highlight{\g{2.11}{0.34}}
        & & \g{0.76}{0.01}
        \\
        
        DQN 
        & & \g{0.32}{0.14} 
        & & \g{54.8}{10.3} 
        & & \g{0.81}{0.05} 
        & & \g{1.06}{0.11} 
        & & \g{0.04}{0.02}
        \\
        
        SAC 
        & & \g{0.65}{0.04} 
        & & \g{84.7}{1.5}  
        & & \g{0.93}{0.02} 
        & & \g{0.54}{0.06} 
        & & \g{0.67}{0.09} 
        \\
        
        PPO
        & & \highlight{\g{0.80}{0.04}}
        & & \highlight{\g{88.3}{0.7}}
        & & \g{0.99}{0.00}
        & & \g{0.12}{0.07} 
        & & \g{0.05}{0.06} 
        \\
        
        PPO-RND 
        & & \g{0.58}{0.04} 
        & & \g{80.7}{1.8}
        & & \g{0.85}{0.08} 
        & & \g{0.83}{0.13} 
        & & \g{0.60}{0.01} 
        \\
        
        Rainbow 
        & & \g{0.51}{0.08} 
        & & \g{72.3}{7.6}
        & & \g{0.69}{0.09} 
        & & \g{1.39}{0.50} 
        & & \g{0.48}{0.02} 
        \\

        GFN $(\beta = 6)$ 
        & & \g{0.36}{0.01} 
        & & \g{57.1}{1.0}
        & & \highlight{\g{0.51}{0.06}} 
        & & \g{1.94}{0.30} 
        & & \highlight{\g{0.85}{0.02}}
        \\

        GFN $(\beta = 32)$ 
        & & \g{0.56}{0.01} 
        & & \g{77.1}{1.1}
        & & \g{0.80}{0.04}
        & & \g{0.99}{0.24} 
        & & \g{0.82}{0.00}
        \\

        GFN $(\beta = 64)$
        & & \g{0.61}{0.00} 
        & & \g{81.9}{0.6}
        & & \g{0.93}{0.02}
        & & \g{0.58}{0.10} 
        & & \g{0.79}{0.02}
        \\
        
        \bottomrule
    \end{tabular}
    \end{adjustbox}
    \label{tab:pf-inf-50}
\end{table}

%% file: tables/oracle-inf-100-table.tex
\setlength\tabcolsep{5pt}
\begin{table}[!h]
    \centering
    \caption{\textbf{Optimizing on pTM scores from {\color{red!70}{ESMFold}} (Sequence length $100$)}: Final scores of the generated protein sequences with a budget of $10 \times 10^7$ oracle reward evaluations. \emph{Best} protein confidence scores and \emph{second best} diversity scores are highlighted. All scores are computed on the outputs of the 3B ESMFold model.}
    \begin{adjustbox}{center,scale=0.9}
    \begin{tabular}{lccccccccccc}
        \toprule
         & & \multicolumn{3}{c}{\textbf{Protein Confidence}} 
         & & \multicolumn{5}{c}{\textbf{Diversity Score}} \\
        \cmidrule(lr){3-5}
        \cmidrule(lr){7-11}
        & & pTM ($\uparrow$) 
         & & pLDDT ($\uparrow$) 
         & & MP-TM ($\downarrow$) 
         & & MP-RMSD ($\uparrow$) 
         & & MP-HD ($\uparrow$)  \\
        \midrule

        Random 
        & & \g{0.40}{0.00} 
        & & \g{46.9}{0.48} 
        & & \g{0.35}{0.01} 
        & & \g{3.99}{0.02} 
        & & \g{0.90}{0.02}
        \\

        MH-MCMC 
        & & \oraclehighlight{\g{0.93}{0.00}} 
        & & \oraclehighlight{\g{92.2}{1.09}}
        & & \g{1.00}{0.00} 
        & & \g{0.04}{0.03} 
        & & \g{0.01}{0.01} 
        \\

        DQN 
        & & \g{0.37}{0.03} 
        & & \g{42.2}{2.74} 
        & & \g{0.62}{0.14} 
        & & \g{2.76}{0.75} 
        & & \g{0.15}{0.12} 
        \\

        SAC
        & & \g{0.71}{0.02} 
        & & \g{78.5}{2.32} 
        & & \g{0.71}{0.12} 
        & & \g{1.87}{0.43} 
        & & \g{0.66}{0.06} 
        \\

        PPO
        & & \g{0.86}{0.02} 
        & & \g{88.9}{1.90} 
        & & \g{1.00}{0.00} 
        & & \g{0.03}{0.03} 
        & & \g{0.01}{0.01} 
        \\

        PPO-RND 
        & & \g{0.50}{0.05} 
        & & \g{55.4}{5.16} 
        & & \g{0.91}{0.11} 
        & & \g{0.53}{0.64} 
        & & \g{0.16}{0.20} 
        \\

        Rainbow
        & & \g{0.45}{0.06} 
        & & \g{52.7}{3.70} 
        & & \g{1.00}{0.00} 
        & & \g{0.00}{0.00} 
        & & \g{0.00}{0.00} 
        \\

        GFN $(\beta = 6)$ 
        & & \g{0.42}{0.03} 
        & & \g{48.6}{1.13} 
        & & \oraclehighlight{\g{0.35}{0.01}}
        & & \oraclehighlight{\g{3.99}{0.11}}
        & & \oraclehighlight{\g{0.90}{0.01}}
        \\

        GFN $(\beta = 32)$ 
        & & \g{0.46}{0.01}
        & & \g{53.7}{1.39}
        & & \g{0.38}{0.01}
        & & \g{3.59}{0.10}
        & & \g{0.88}{0.02}
        \\

        GFN $(\beta = 64)$
        & & \g{0.55}{0.04}
        & & \g{63.6}{1.87}
        & & \g{0.55}{0.03}
        & & \g{2.66}{0.33}
        & & \g{0.78}{0.07}
        \\
        
        \bottomrule
    \end{tabular}
    \end{adjustbox}
    \label{tab:oracle-inf-100}
\end{table}

%% file: tables/oracle-inf-150-table.tex
\setlength\tabcolsep{5pt}
\begin{table}[!h]
    \centering
    \caption{\textbf{Optimizing on pTM scores from {\color{red!70}{ESMFold}} (Sequence length $150$)}: Final scores of the generated protein sequences with a budget of $10 \times 10^7$ oracle reward evaluations. \emph{Best} protein confidence scores and diversity scores are highlighted. All scores are computed on the outputs of the 3B ESMFold model.}
    \begin{adjustbox}{center,scale=0.9}
    \begin{tabular}{lccccccccccc}
        \toprule
         & & \multicolumn{3}{c}{\textbf{Protein Confidence}} 
         & & \multicolumn{5}{c}{\textbf{Diversity Score}} \\
        \cmidrule(lr){3-5}
        \cmidrule(lr){7-11}
        & & pTM ($\uparrow$) 
         & & pLDDT ($\uparrow$) 
         & & MP-TM ($\downarrow$) 
         & & MP-RMSD ($\uparrow$) 
         & & MP-HD ($\uparrow$)  \\
        \midrule

        % Random 
        % & & \g{0.40}{0.00} 
        % & & \g{46.9}{0.48} 
        % & & \g{0.35}{0.01} 
        % & & \g{3.99}{0.02} 
        % & & \g{0.89}{0.02}
        % \\

        MH-MCMC 
        & & \oraclehighlight{\g{0.93}{0.00}} 
        & & \oraclehighlight{\g{89.77}{0.60}}
        & & \g{1.00}{0.00} 
        & & \g{0.03}{0.02} 
        & & \g{0.01}{0.00} 
        \\

        % DQN 
        % & & \g{0.37}{0.03} 
        % & & \g{42.2}{2.74} 
        % & & \g{0.62}{0.14} 
        % & & \g{2.76}{0.75} 
        % & & \g{0.15}{0.12} 
        % \\

        SAC
        & & \g{0.36}{0.02} 
        & & \g{38.22}{0.93} 
        & & \oraclehighlight{\g{0.34}{0.01}}
        & & \oraclehighlight{\g{4.81}{0.12}}
        & & \oraclehighlight{\g{0.87}{0.02}}
        \\

        PPO
        & & \g{0.91}{0.01} 
        & & \g{87.64}{1.55} 
        & & \g{1.00}{0.00} 
        & & \g{0.11}{0.04} 
        & & \g{0.03}{0.01} 
        \\

        PPO-RND 
        & & \g{0.39}{0.02} 
        & & \g{43.65}{5.36} 
        & & \g{0.83}{0.01} 
        & & \g{1.24}{0.18} 
        & & \g{0.24}{0.03} 
        \\

        % Rainbow
        % & & \g{0.45}{0.06} 
        % & & \g{52.7}{3.70} 
        % & & \g{1.00}{0.00} 
        % & & \g{0.00}{0.00} 
        % & & \g{0.00}{0.00} 
        % \\

        GFN $(\beta = 6)$ 
        & & \g{0.35}{0.01} 
        & & \g{37.70}{1.96}
        & & \oraclehighlight{\g{0.34}{0.02}}
        & & \g{4.74}{0.21}
        & & \g{0.86}{0.03}
        \\

        GFN $(\beta = 32)$ 
        & & \g{0.36}{0.01}
        & & \g{39.97}{2.23}
        & & \g{0.35}{0.03}
        & & \g{4.67}{0.25}
        & & \oraclehighlight{\g{0.87}{0.02}}
        \\

        GFN $(\beta = 64)$
        & & \g{0.40}{0.03}
        & & \g{42.04}{1.07}
        & & \g{0.62}{0.01}
        & & \g{2.62}{2.15}
        & & \g{0.52}{0.46}
        \\
        
        \bottomrule
    \end{tabular}
    \end{adjustbox}
    \label{tab:oracle-inf-150}
\end{table}

%% file: tables/oracle-inf-50-biometrics.tex
\setlength\tabcolsep{5pt}
\begin{table}[h]
    \centering
    \caption{Biophysical metrics of $50$-residue proteins generated by optimizing pTM scores using \textbf{{\color{red!70}{ESMFold}}}. The last column is the Distributional Conformity Score comparing generated proteins to reference proteins of the same length in PDB.}
    
    \begin{adjustbox}{center,scale=0.9}
    \begin{tabular}{lccccccccccc}
        \toprule
         % & & \multicolumn{3}{c}{\textbf{Protein Confidence}} 
         % & & \multicolumn{5}{c}{\textbf{Diversity Score}} \\
        % \cmidrule(lr){3-5}
        % \cmidrule(lr){7-11}
         & & $W_{mol}$ 
         & & Instability Index 
         & & Isoelectric Point (pI)
         & & GRAVY 
         & & DCS ($\uparrow$)\\
         \midrule

        % Random 
        % & & $6087.23$
        % & & $48.47$
        % & & $7.63$
        % & & $-0.52$ 
        % & & 
        % \\

        MH-MCMC 
        & & $5598.95$
        & & $35.90$
        & & $5.87$
        & & $0.22$
        & & \underline{$0.53$}
        \\

        DQN 
        & & $5990.59$
        & & $41.57$
        & & $7.69$
        & & $-0.16$
        & & $0.31$
        \\

        SAC
        & & $6066.13$
        & & $47.09$
        & & $7.75$
        & & $0.30$
        & & $0.39$
        \\

        PPO
        & & $5955.30$
        & & $49.38$
        & & $6.33$
        & & $0.57$
        & & \oraclehighlight{$0.56$}
        \\

        PPO-RND 
        & & $6078.18$
        & & $54.13$
        & & $8.11$
        & & $0.18$
        & & $0.41$
        \\

        Rainbow
        & & $5908.24$
        & & $41.17$
        & & $6.61$
        & & $0.37$
        & & $0.47$
        \\

        GFN $(\beta = 6)$ 
        & & $6166.45$
        & & $44.34$
        & & $8.04$
        & & $-0.36$
        & & $0.29$
        \\

        GFN $(\beta = 32)$ 
        & & $6106.51$
        & & $51.47$
        & & $9.18$
        & & $-0.08$
        & & $0.41$
        \\

        GFN $(\beta = 64)$
        & & $6071.02$
        & & $54.17$
        & & $8.25$
        & & $-0.29$
        & & $0.41$
        \\
        
        \bottomrule
    \end{tabular}
    \end{adjustbox}
    \label{tab:oracle-inf-50-biometrics}
\end{table}

%% file: tables/oracle-inf-100-biometrics.tex
\setlength\tabcolsep{5pt}
\begin{table}[h]
    \centering
    \caption{Biophysical metrics of $100$-residue proteins generated by optimizing pTM scores using \textbf{{\color{red!70}{ESMFold}}}. The last column is the Distributional Conformity Score comparing generated proteins to reference proteins of the same length in PDB.}
    
    \begin{adjustbox}{center,scale=0.9}
    \begin{tabular}{lccccccccccc}
        \toprule
         % & & \multicolumn{3}{c}{\textbf{Protein Confidence}} 
         % & & \multicolumn{5}{c}{\textbf{Diversity Score}} \\
        % \cmidrule(lr){3-5}
        % \cmidrule(lr){7-11}
         & & $W_{mol}$ 
         & & Instability Index 
         & & Isoelectric Point  (pI)
         & & GRAVY 
         & & DCS ($\uparrow$)\\
         \midrule

        % Random 
        % & & $11888.94$
        % & & $47.29$
        % & & $6.44$
        % & & $-0.42$
        % \\

        MH-MCMC 
        & & $11470.28$
        & & $35.67$
        & & $5.52$
        & & $-0.21$
        & & \oraclehighlight{$0.53$}
        \\

        DQN 
        & & $12096.87$
        & & $43.84$
        & & $8.74$
        & & $-0.54$
        & & $0.29$
        \\

        SAC
        & & $12079.58$
        & & $46.09$
        & & $6.78$
        & & $0.87$
        & & $0.27$
        \\

        PPO
        & & $11697.12$
        & & $34.69$
        & & $10.55$
        & & $0.36$
        & & \underline{$0.52$}
        \\

        PPO-RND 
        & & $12009.17$
        & & $51.34$
        & & $7.10$
        & & $-0.52$
        & & $0.33$
        \\

        Rainbow
        & & $12121.63$
        & & $51.81$
        & & $9.25$
        & & $-0.33$
        & & $0.17$
        \\

        GFN $(\beta = 6)$ 
        & & $12058.39$
        & & $44.49$
        & & $7.55$
        & & $-0.46$
        & & $0.34$
        \\

        GFN $(\beta = 32)$ 
        & & $12091.40$
        & & $44.00$
        & & $7.47$
        & & $-0.35$
        & & $0.29$
        \\

        GFN $(\beta = 64)$
        & & $12228.52$
        & & $45.64$
        & & $8.23$
        & & $-0.14$
        & & $0.18$
        \\
        
        \bottomrule
    \end{tabular}
    \end{adjustbox}
    \label{tab:oracle-inf-100-biometrics}
\end{table}

%% file: tables/oracle-inf-150-biometrics.tex
\setlength\tabcolsep{5pt}
\begin{table}[h]
    \centering
    \caption{Biophysical metrics of $150$-residue proteins generated by optimizing pTM scores using \textbf{{\color{red!70}{ESMFold}}}. The last column is the Distributional Conformity Score comparing generated proteins to reference proteins of the same length in PDB.}
    
    \begin{adjustbox}{center,scale=0.9}
    \begin{tabular}{lccccccccccc}
        \toprule
         % & & \multicolumn{3}{c}{\textbf{Protein Confidence}} 
         % & & \multicolumn{5}{c}{\textbf{Diversity Score}} \\
        % \cmidrule(lr){3-5}
        % \cmidrule(lr){7-11}
         & & $W_{mol}$ 
         & & Instability Index 
         & & Isoelectric Point  (pI)
         & & GRAVY 
         & & DCS ($\uparrow$)\\
         \midrule

        MH-MCMC 
        & & $17818.09$
        & & $32.19$
        & & $7.99$
        & & $-0.16$
        & & $0.09$
        \\

        SAC
        & & $17841.44$
        & & $41.99$
        & & $7.17$
        & & $-0.47$
        & & \oraclehighlight{$0.22$}
        \\

        PPO
        & & $17971.68$
        & & $48.31$
        & & $7.53$
        & & $0.09$
        & & $0.03$
        \\

        PPO-RND 
        & & $17920.13$
        & & $50.46$
        & & $8.87$
        & & $-0.37$
        & & \underline{$0.19$}
        \\

        GFN $(\beta = 6)$ 
        & & $18031.62$
        & & $41.95$
        & & $7.15$
        & & $-0.45$
        & & $0.16$
        \\

        GFN $(\beta = 32)$ 
        & & $18008.17$
        & & $46.20$
        & & $7.49$
        & & $-0.39$
        & & $0.15$
        \\

        GFN $(\beta = 64)$
        & & $18129.90$
        & & $38.15$
        & & $7.19$
        & & $-0.53$
        & & $0.09$
        \\
        
        \bottomrule
    \end{tabular}
    \end{adjustbox}
    \label{tab:oracle-inf-150-biometrics}
\end{table}

%% file: tables/pf-dcs.tex
\setlength\tabcolsep{5pt}
\begin{table}[t]
    \centering
    \caption{\textbf{Optimizing on pTM scores from finetuned {\color{blue!70}{proxy}} model (ESM-PF)}: Distributional Conformity Score}
    \begin{tabular}{lcc}
        \toprule
         & \multicolumn{2}{c}{DCS $\uparrow$} \\
          \cmidrule(lr){2-3}
         & $L=50$          & $L=100$          \\
         \midrule
PPO      & 0.3074          & 0.1163           \\
SAC      & 0.3361          & 0.3280           \\
DQN      & \highlight{0.4807}          & 0.1397           \\
PPO-RND  & 0.3268          & 0.0457           \\
Rainbow  & 0.4340          & \highlight{0.6140}           \\
GFN $(\beta = 6)$  & 0.3622          & 0.2395           \\
GFN $(\beta = 32)$ & 0.3667          & 0.3803           \\
GFN $(\beta = 64)$ & 0.2908          & 0.2376           \\
MH-MCMC  & 0.2583          & 0.0321           \\
    \bottomrule
    \end{tabular}
    \label{tab:pf-dcs}
\end{table}

%% file: tables/oracle-aa_dist-mae.tex
\setlength\tabcolsep{5pt}
\begin{table}[t]
    \centering
    \caption{\textbf{Optimizing on pTM scores from {\color{red!70}{ESMFold}}}: Mean absolute error in normalized frequency of amino acids between proposed sequences and naturally occurring proteins from the PDB.}
    \begin{tabular}{lccc}
        \toprule
         & \multicolumn{3}{c}{MAE $\downarrow$} \\
          \cmidrule(lr){2-4}
         & $L=50$     & $L=100$    & $L=150$    \\
         \midrule
PPO      & 0.047     & 0.029     & 0.025     \\
SAC      & 0.036     & 0.036     & 0.021     \\
DQN      & 0.027     & 0.022     & \oraclehighlight{0.019}     \\
PPO-RND  & 0.032     & 0.017     & 0.021     \\
Rainbow  & 0.024     & 0.018     & 0.050     \\
GFN $(\beta = 6)$  & \oraclehighlight{0.021}     & \oraclehighlight{0.016}     & 0.023     \\
GFN $(\beta = 32)$ & 0.025     & 0.018     & 0.022     \\
GFN $(\beta = 64)$ & 0.026     & 0.021    & 0.021     \\
MH-MCMC  & 0.031     & 0.017     & 0.023    \\
        \bottomrule
    \end{tabular}
    \label{tab:oracle-aa_dist-mae}
\end{table}

%% file: tables/pf-50-mae.tex
\setlength\tabcolsep{5pt}
\begin{table}[t]
    \centering
    \caption{\textbf{Optimizing on pTM scores from finetuned {\color{blue!70}{proxy}} model (ESM-PF)}: Mean absolute error in normalized frequency of amino acids between proposed sequences and naturally occurring proteins from PDB}
    \begin{tabular}{lcc}
        \toprule
         & \multicolumn{2}{c}{MAE $\downarrow$} \\
          \cmidrule(lr){2-3}
         & $L=50$           & $L=100$           \\
         \midrule
PPO      & 0.0526           & 0.0285            \\
SAC      & 0.0462           & 0.0390            \\
DQN      & 0.0207           & 0.0230            \\
PPO-RND  & 0.0307           & 0.0204            \\
Rainbow  & 0.0321           & 0.0265            \\
GFN $(\beta = 6)$  & 0.0228           & \highlight{0.0195}            \\
GFN $(\beta = 32)$ & 0.0275           & 0.0233            \\
GFN $(\beta = 64)$ & 0.0363           & 0.0212            \\
MH-MCMC  & \highlight{0.0203}           & 0.0240            \\
    \bottomrule
    \end{tabular}
    % \end{adjustbox}
    \label{tab:pf-mae}
\end{table}